\documentclass[letterpaper, 10 pt, conference]{ieeeconf}  

\usepackage{setspace}
\usepackage{algorithm}
\usepackage{titlesec}
\usepackage[linguistics]{forest}
\usepackage{tcolorbox}
\usepackage{listings}
\usepackage{adjustbox}
\usepackage{mathtools}
\usepackage{algpseudocode}
\usepackage[hyphens]{url}
\usepackage{lipsum}
\usepackage{graphicx}
\usepackage[utf8]{inputenc}
\IEEEoverridecommandlockouts                              

\title{\LARGE \bf
A Non-linear MPC Local Planner for Tractor-Trailer Vehicles in Forward and Backward Maneuvering
}

\author{Behnam Moradi$^{1}$, Mehran Mehrandezh$^{2}$
\thanks{$^{1}$ University of Regina, Faculty of Engineering and Applied Science, 
        {\tt\small bmn891@uregina.ca}}
\thanks{$^{2}$ University of Regina, Faculty of Engineering and Applied Science
        {\tt\small mehran.mehrandezh@uregina.ca}}
}

\begin{document}

\maketitle
\thispagestyle{empty}
\pagestyle{empty}

\begin{abstract}

Designing a local planner to control tractor-trailer vehicles in forward and backward maneuvering is a challenging control problem in the research community of autonomous driving systems. Considering a critical situation in the stability of tractor-trailer systems, a practical and novel approach is presented to design a non-linear MPC (NMPC) local planner for tractor-trailer autonomous vehicles in both forward and backward maneuvering. The tractor’s velocity and steering angle are considered as control variables. The proposed NMPC local planner is designed to handle jackknife situations, avoiding multiple static obstacles, and path following in both forward and backward maneuvering. The above-mentioned challenges are converted into a constrained problem that can be handled simultaneously by the proposed NMPC local planner. the direct multiple shooting approach is used to convert the optimal control problem (OCP) into a non-linear programming problem (NLP) that can be solved by IPOPT solvers in CasADi. The controller performance is evaluated through different backup and forward maneuvering scenarios in the Gazebo simulation environment in real-time and achieves asymptotic stability in avoiding static obstacles and accurate tracking performance while respecting path constraints. Finally, the proposed NMPC local planner is integrated with an open-source autonomous driving software stack called Autoware.ai.

\end{abstract}

\section{INTRODUCTION}
Given the hard constraints on both state and control variables, the MPC controller is widely used to handle the constrained local planning problem. MPC controller is able to stabilize complex dynamic multi-variable systems while accounting for actuator limitations and system parameters at each sampling time. Therefore, in the case of a local planning problem, the MPC controller basically solves a nonlinear optimization problem at each sampling time inside a finite horizon window. MPC controllers are also utilized as optimal local planners in mobile robots to handle trajectory following problems.\cite{worthmann2015model}.

Mehrez et al \cite{worthmann2015model} developed a stabilizing MPC controller to control the  unicycle nonholonomic mobile robots. This model is later applied to solve trajectory planning and optimal waypoints following while minimizing traveling distance and fuel consumption in the optimization problem. Considering static and dynamic obstacles in the operation environment, Tallamraju et al \cite{lu2022real} designed an MPC controller to handle obstacle avoidance in multi-robot target tracking scenarios. Introducing static and dynamic obstacles into the system’s constraints usually puts the solver in the local minima trap. In their proposed approach, three methods are introduced to resolve this issue and facilitate convergence to  the desired state. They introduced an MPC local planner which enables a UAV to avoid dynamic incoming obstacles  with perception constraints. They combined  the motion and perception objectives in order to find a feasible and collision-free trajectory while considering the limited FOV. This approach was later extended to optimize an additional perception objective which maximizes the visibility of the target point in the RGB data.

Yang et al \cite{bin2012constrained} proposed an MPC controller to enable precise path tracking for a backing-up tractor-trailer system with off-axle hitching. To begin, they presented an MLD model to characterize the tractor-trailer's open-loop kinematics. An expanded cost function given in the quadratic form is addressed for the resulting MLD model. The primary idea behind the suggested cost function is to take into account both the tractor's steering angle and the trailer's orientation error at the same time. An equivalent linear quadratic tracking issue is described to minimize the tracking error of the trailer's location and orientation. Then, using finite-time optimal controllers, a receding horizon finite-time optimum controller is built. To address the jackknife avoidance problem in tractor-trailer systems, Minglei et al  \cite{fan2019anti} proposed a jackknife avoidance trajectory tracking control technique for tractor-trailer vehicles in the process of reversing perpendicular parking using constrained Model Predictive Control (MPC). First, they started by studying the influencing variables of the size and stability of the vehicle hitch angle, then a critical hitch angle is established for jackknife prevention and collision-free constraint. As a result, the issues of avoiding collision and jackknifing are redefined as constraint problems. As a result, the MPC algorithm, which has the advantage of being able to handle many constraints problems, can be used to solve the jackknife and tractor-trailer collision problems. Finally, simulation results reveal that the suggested control strategy functions as expected during this complicated vehicle system's reversing operation.

To address the jackknife problem at low speed, \cite{beglini2020anti} offered a control approach for backward maneuvering, in which jackknifing can occur at low speeds, that drives the vehicle along a reference Cartesian trajectory while avoiding hitch angle divergence. A feedback control law is proposed by combining two actions: a tracking term computed using input-output linearization and a correction term generated using ISMPC, an inherently stable MPC technique used for stable inversion of nonminimum-phase systems. They verified the proposed method using simulation and a real tractor-trailer vehicle.

MPC local planners are also utilized in highway driving. Van et al \cite{van2015real} proposed a finite-horizon optimum control problem (OCP) that combines a prediction model in spatial coordinates for the vehicle and surrounding traffic in order to consider all relevant information for the road and surrounding traffic. This permits road features (such as curvature) to appear in the prediction horizon as known variables. The resulting constrained nonlinear least-squares problem's objective function strikes a balance between tracking performance, driver comfort, and maintaining a safe distance from other road users. The OCP is solved using a direct multiple shooting solution approach implemented in a real-time iteration scheme utilizing ACADO code generation, and the results are compared to a feedback strategy that uses the interior point solver IPOPT to solve the complete nonlinear program in each time step.

MPC local planners can also be used to handle path-tracking problems in presence of nonholonomic limitations, unpredictable disruption, and other physical limits. Ming et al \cite{yue2017composite} proposed a combination of two powerful control approaches to offer a composite path-tracking control strategy to address various difficulties emerging from vehicle kinematic and dynamic levels. The proposed composite control structure consists of a model predictive control (MPC)-based posture controller and a direct adaptive fuzzy-based dynamic controller. The former posture controller can make the underactuated trailer midpoint follow an arbitrary reference trajectory determined by the earth-fixed frame while also meeting certain physical constraints. Meanwhile, the latter dynamic controller allows the vehicle velocities to track the desired velocities created by the former, and the dynamic controller's global asymptotical convergence.

MPC is a model-based local planner. The performance of the controller depends on the solver's ability and the way its model parameters are defined. In the case of a tractor-trailer vehicle, the kinematic model of the system is non-linear and consequently, the MPC local planner will be designed as a non-linear controller. Taking into account the articulated vehicle's mechanical and physical constraints, as well as its full kinematic model, Thaker et al \cite{nayl2015effect} proposed an MPC-based path planner which is part of the Bug-Like algorithm family \cite{nayl2015real}.  The proposed MPC local planner was able to regulate the vehicle's lateral motion during the online motion planning process. Their proposed integrated path planning and control technique was examined for efficiency and sensitivity in a variety of simulated test cases, with the dependence on the specified kinematic parameters highlighted.

MPC local planners are also utilized to handle the path-planning problem for a tractor with multiple trailers connected to it. Mohamed et al \cite{mohamed2018optimal} proposed a combination of the artificial potential field (APF) method and optimal control theory to design a local planner to handle the path planning mission for an autonomous truck with two trailers in autonomous navigation.  The optimal control theory is used to design an optimal free-obstacle path for the robotic vehicle from a starting point to the objective location. The obstacle-avoidance strategy is mathematically represented by a potential function based on the proposed sigmoid function. The potential field model that was created can provide an accurate analytic description of three-dimensional objects.

In this article, a full-scale solution is introduced to solve the optimal control theory by transforming it into a non-linear programming problem and solving it with the Multiple Shooting method using the IPOPT  solver. Our first task is to design and implement the local planner in a simple python-based simulation. Then we will be designing a full-scale simulation environment in Gazebo-ROS \cite{bingham2019toward} which has the ability to use a physics engine. Our NMPC local planner will be implemented on a simulated tractor-trailer system and it will be later integrated with one of the world-wide used an open-sourced autonomous driving platform called Autoware.ai \cite{Autoware2013}. During the simulation, we will be constructing a 3D HD-Map of the environment. The map is based on point-cloud data generated by a Velodyne VLP-16 lidar. We will be implementing NDT mapping algorithm to construct the HD-Map and then NDT-matching to localize the vehicle in the generated HD-Map. Therefore, we will be providing our NMPC local planner with real-time feedback on the current position of the vehicle. Then the planner will output the linear velocity and steering angle in real-time during path tracking and obstacle avoidance mission. 

Our NMPC local planner is able to handle path tracking and static obstacle avoidance missions in both forward and backward maneuvering while respecting actuator limitations and physical constraints of the environment. The performance of our local planner is reported by calculating path tracking errors and execution time during every forward and backward maneuvering mission.

\section{Kinematic Modeling}
The kinematic model of our standard tractor-trailer system is extracted based on the following assumptions.
\begin{enumerate}
  \item The tractor's steering system is designed based on the Ackermann mechanism.
  \item The surface of the road is flat and smooth. Therefore, we can ignore suspension dynamics, roll motions, and actuator dynamics. 
  \item The system is only allowed to operate at low speed (up to 0.2 m/s in backward maneuvering) so that the trailer does not experience lateral slippage.
  \item The trailer's kingpin is directly connected to the tractor's middle rear axle, which is called the hitch point of the tractor-trailer system.
\end{enumerate}

\begin{figure}[h!]
  \centering
  \includegraphics[scale=0.5]{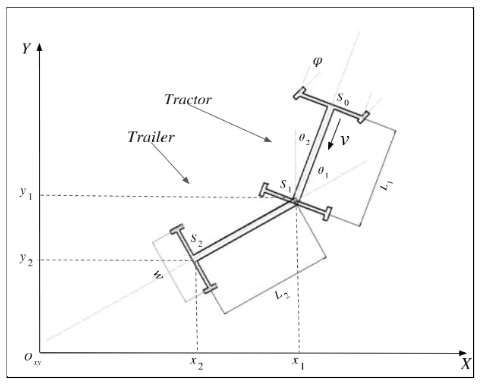}
  \caption{The simplified schematics of the tractor-trailer system. $O_{xy}$ is considered to be the origin of the operating field.}
  \label{kinematic_shape}
\end{figure}

We consider $O_{xy}$ as the reference frame of the system. $\theta_1$ is the orientation of the tractor with respect to the trailer. $\theta_2$ is the orientation of the tractor with respect to the vertical axis (y-axis). The width of both the tractor and the trailer is denoted by $w$. $\varphi$ is the steering angle of the tractor with respect to the tractor’s main chassis. The velocity of the trailer is denoted by $v$ which is measured at point $S_2$. $l_1$ and $l_2$ indicate the length of the tractor and the trailer, respectively. The coordinates of $S_1$ and $S_2$ points are denoted by $[x_1, y_1]$ and $[x_2, y_2]$, respectively. Based on these declarations, The following equations are extracted to describe the kinematic relationship between the tractor and the trailer in both forward and backward maneuvering.

\begin{equation}
    \frac{dx_2}{dt} = v cos(\theta_1)sin(\theta_2)
\end{equation}
\begin{equation}
    \frac{dy_2}{dt} = -v cos(\theta_1)cos(\theta_2)
\end{equation}
\begin{equation}
    \frac{d\theta_1}{dt} = v\frac{sin(\theta_1) + tan(\varphi)}{L_2}
\end{equation}
\begin{equation}
    \frac{d\theta_2}{dt} = v\frac{sin(\theta_1)}{L_2}
\end{equation}

\section{Proposed NMPC Local Planner}

MPC, as an optimal controller, is able to return control actions by solving an optimal control problem in a finite control horizon. Therefore, control actions are calculated sequentially over a predefined finite horizon. At each sampling time, only the first control action is applied to the system’s actuators. The process of generating sequential control actions takes place at each sampling time until the terminal condition is met. The feedback control system is then established using the concept of the receding horizon strategy. MPC is able to control nonlinear multi input multi output (MIMO) systems while accounting for states and actuator limitations. Therefore, MPC has been found to be extremely useful in a wide range of industrial applications.

As we mentioned before, in MPC, the control actions (tractor’s velocity and steering angle) are the minimizers to minimize the objective function. Therefore, they can be obtained by solving an open-loop optimal control problem at each sampling time. The open-loop sequential control actions are returned over the prediction horizon $N_c \in N$.

\begin{equation}
    u = (u(0), u(1), ..., u(N-1)) \in U^{N_c}
    \label{eq:control_actions}
\end{equation}

Therefore, the online optimal control problem can be formulated as the following:

\begin{equation}
    min \qquad J_{N_c}(x_0, u)|u \in R^{{n_u}{N_c}}
\end{equation}

Subject to:

\begin{equation}
    x(0) = x_0
\end{equation}

\begin{equation}
    x(k+1)=f(x(k), u(k)) \qquad \forall k \in {0, 1, ..., N_c - 1}
\end{equation}

\begin{equation}
    x(k) \in X \qquad \forall k \in {1, ..., N_c - 1}
\end{equation}

\begin{equation}
    u(k) \in U \qquad \forall k \in {0, 1, ..., N_c - 1}
\end{equation}

The above-mentioned objective function is a summation of running cost and terminal cost which is formulated as Equations \ref{eq:running_cost} and \ref{eq:terminal_cost}:

\begin{equation}
    \sum_{k=0}^{N_c - 1} {l(x(k), u(k))} 
    \label{eq:running_cost}
\end{equation}

\begin{equation}
    F(x(N_c))https://www.overleaf.com/project/62c0cb540ebaa687dffdd958
    \label{eq:terminal_cost}
\end{equation}

The objective function is formulated as Equation \ref{eq:obj_func_2}:

\begin{equation}
    J_{N_c}(x_0, u) =  \sum_{k=0}^{N_c-1}{l(x(k), u(k)) + F(x(N_c))}
    \label{eq:obj_func_2}
\end{equation}

Both system's states and control actions are penalized inside the running cost equation. At each sampling time $k$, the system's states $x(k)$ are predicted and control actions $u(k)$ is obtained over the prediction horizon $N$. The running cost is calculated by penalizing the followings:

\begin{itemize}
    \item The difference between the current predicted and the reference state
    \item The difference between the computed and reference control actions
\end{itemize}

The terminal cost is also the difference between the last point of the predicted trajectory and its reference point. Therefore, the running cost can be written in the following form:

\begin{equation}
    l(x(k), u(k)) = ||x(k) - x^r||^2_Q + ||u(k) - u^r||^2_{R^2}
    \label{eq:running_cost_2}
\end{equation}

$Q$ and $R$ are the matrices used to penalize the states and control actions. Therefore, they should be symmetric and positive definite. The main idea behind penalizing the control actions is to avoid expensive control actions in terms of energy consumption. The terminal cost can be written as the following:

\begin{equation}
    F(x(N_c)) = ||x(N_c) - x^r(N_c)||^2_P
    \label{eq:terminal_cost_2}
\end{equation}

Once the objective function (i.e. the summation of running and terminal costs) is solved and the termination condition is met, a sequence of control actions over the prediction horizon is returned by the $IPOPT$ solver. This control sequence is called the Minimizer Vector:

\begin{equation}
    u = (u(0), u(1), ..., u(N-1)) \in U^{N_c}
    \label{eq:control_actions}
\end{equation}

As we discussed at the beginning of this section, an optimal control problem is usually subjected to constraints on the followings:

\begin{itemize}
    \item System’s states, e.g. $X$
    \item Control actions, e.g. $U$
    \item Terminal constraints (equality and inequality constraints)
\end{itemize}

In the process of evaluating an MPC controller, feasibility, real-time applicability, and stability are the most important metrics. Thus, the above-mentioned constraints play an important role in this game. State constraints are considered to be soft constraints. Meaning that they can be manipulated by adding a predefined small real number that can be intuitively justified from a feasibility perspective. Therefore, relaxing state constraints not only helps the system to achieve a stable behavior but it helps to maintain feasibility while the system is facing disturbance and noise. The constraints on the control actions, on the other side, are considered to be hard constraints. It means the controller is not allowed to relax them due to the physical limitations of the electrical and mechanical actuators of the system. To guarantee the stability of closed-loop MPC, it is necessary to consider both equality and inequality terminal constraints. In terminal equality constraints, the very last predicted state of the system needs to be equal to its correspondence reference while in terminal inequality constraints, the last predicted state needs to be in a predefined area around the reference state. Terminal constraints can be formulated as Equations \ref{eq:terminal_equality_cons} and \ref{eq:terminal_inequality_cons}:

\begin{equation}
    x(N_c) - x^r(N_c) = 0
    \label{eq:terminal_equality_cons}
\end{equation}

\begin{equation}
    x(N_c) \in A(x^r(N_c))
    \label{eq:terminal_inequality_cons}
\end{equation}

where  $x(N_c)$ and $x_r(N_c)$ indicate the last state vector and its reference value, respectively,  at the end of the prediction horizon $N_c$. A simplified step-by-step implementation of nonlinear closed-loop MPC is presented by Algorithm 1.

\begin{algorithm}
\caption{Algorithmic implementation of NMPC controller}\label{alg:nmpc-controller}
\begin{algorithmic}
\State \textbf{Input 1:} Initial pose $x_0$
\State \textbf{Input 2:} Target pose $x_s$
\State \textbf{Input 3:} NMPC parameters

\While{\texttt{$Error > min-allowable-error$}}
\State - Calculate the Error: $Error = norm(x_s - x_0)$
\State - Read the states feedback : $ \widehat{x} := x(p) \in X$
\State - Set the first state vector: $x(0) = x(p)$
\State - Solve the OCP and return the sequence 
\State \, of control actions over the prediction 
\State \, horizon: $u = {u(0, u(1), ..., U(N-1))} \in U^N_c$
\State - Closed-loop control: Close the feedback loop 
\State \, by applying the first control action to the 
\State \, system’s actuator(s)
\EndWhile

\end{algorithmic}
\end{algorithm}

\subsection{Online Optimization}
An infinite optimal control problem (OCP) can be transformed into a non-linear programming problem (NLP) using direct methods. The resulting NLP takes the following general form:

\begin{equation}
    min \, a(w)
    \label{eq:obj_nlp}
\end{equation}

subject to:

\begin{equation}
    b(w) = 0
    \label{eq:nlp_cons_aw}
\end{equation}

\begin{equation}
    c(w) = 0
    \label{eq:nlp_cons_cw}
\end{equation}

where $w$ is the optimization degrees of freedom and all three functions, e.g. objective function $a(w)$, equality constraints $b(w)$, and inequality constraints $c(w)$ are differentiable. Speaking from the technical implementation of direct methods, they parametrize the sequential control actions over the prediction horizon. But it does not mean all direct methods are similar. Rather, they have fundamental differences in the way they handle the states over the prediction horizon. 

Some direct methods use sequential methods in which the state trajectory $x(t)$ is treated as an implicit function of the controls $u(t)$ and the initial value $x_0$ in sequential techniques, such as forward simulation with an ODE solver in direct single shooting \cite{diehl2006fast}. As a result, simulation and optimization iterations go one after another, and the NLP only has discretized sequential control actions as optimization degrees of freedom.

Simultaneous techniques, on the other hand, preserve a state trajectory parameterization as optimization variables within the NLP and add suitable equality constraints representing the ODE model. As a result, simulation and optimization occur simultaneously, with the states representing a valid ODE solution corresponding to the control trajectory only when the NLP is solved.

Direct collocation \cite{stryk1993numerical} and direct multiple shooting \cite{bock2000direct} are the two most prevalent forms of the simultaneous technique.

The direct multiple shooting approach attempts to combine the features of a simultaneous method, such as collocation, with the main benefit of single shooting, notably the ability to apply adaptive and error-controlled ODE solvers. At the very first step, in the multiple shooting method, the control actions are discretized as follows:

\begin{equation}
    u(t) = q_i, \, t \in [t_i, t_{i+1}]
    \label{eq:multiple_shooting_control_actions}
\end{equation}

In the second step, starting with an arbitrary beginning value $s_i$, we solve the ODE on each interval $[t_i, t_{i+1}]$ separately:

\begin{equation}
    \dot{x}_i(t) = f(x_i(t), q_i), \, t \in [t_i, t_{i+1}]
    \label{eq:ode}
\end{equation}

\begin{equation}
    x_i(t_i) = s_i
    \label{eq:ode_x}
\end{equation}

The sequential systems  states (state’s trajectory) $x_i(t_i; q_i, s_i)$ can be obtained by solving these initial value problems numerically. The additional arguments after the semicolon are used to indicate the interval's reliance on its initial values and controls. At the same time, once the ODE solution is obtained, the running cost can be computed as follows:

\begin{equation}
    l_i(s_i, q_i) =\int\limits_{t_i}^{t_i+1} L(x_i(t_i; q_i, s_i),q_i) \ dt
    \label{eq:multiple_shooting_running_cost}
\end{equation}

At the beginning of each interval, to ensure the continuity of the solution, $s_i$ is replaced by its previous value at the previous iteration. It means we use the previous states to initialize the ODE solver at the current state. Therefore, the following NLP formulation can be obtained:

\begin{equation}
    \mathop{min}_{{s, q}} \qquad  \sum_{i=0}^{N-1} l_i(s_i, q_i) + E(s_N)
    \label{eq:multiple_shotting_obj_func}
\end{equation}

Subjected to constraints on the initial value of the system’s states (\ref{eq:initial_value_cons}), continuity constraints (\ref{eq:continuity_cons}), system’s states constraints (\ref{eq:systems_state_cons}), and 

\begin{equation}
    s_0 - x_0 = 0
    \label{eq:initial_value_cons}
\end{equation}

\begin{equation}
    s_{i+1} - x_i(t_{i+1}:q_i, s_i) = 0, \, i \in [0,1,...,N-1]
    \label{eq:continuity_cons}
\end{equation}

\begin{equation}
    h(s_i, q_i) \geq 0, \,  i \in [0,1,...,N-1]
    \label{eq:systems_state_cons}
\end{equation}

\begin{equation}
    r(s_N) = 0
    \label{eq:terminal_const_2}
\end{equation}

Direct multiple shooting is commonly considered a simultaneous approach, like collocation, because it only produces a valid ODE solution after the optimization iterations are completed. However, because it incorporates aspects of both a pure sequential and a pure simultaneous technique, it is frequently referred to as a hybrid method.

\subsection{CasADi: A Software Framework for Online Optimization}
CasADi originated as an algorithmic differentiation (AD) tool with a syntax similar to that of a computer algebra system (CAS). While cutting-edge AD is still a big part of CasADi, in recent years the focus has shifted to optimization. CasADi provides a set of general-purpose building blocks that reduce the time and effort required to design a large number of numerical optimal control problems. 

CasADi is built around a symbolic framework that enables users to create expressions and utilize them to define automatically differentiable functions. These general-purpose expressions are similar to expressions in MATLAB's Symbolic Toolbox or Python's SymPy module in that they have no concept of optimization. Once the expressions have been constructed, they can be utilized to quickly generate additional derivative expressions using AD or evaluated quickly, either in CasADi's virtual machines or by generating self-contained C code using CasADi.

CasADi integrates modeling and mathematical optimization support, just like other algebraic modeling languages. Nonlinear programming Problems (NLPs) and conic optimization problems are two types of optimization problems that are handled. Both linear and quadratic programming problems (QPs) are included in this category. The actual solution is usually done in a derived class, which may employ a CasADi tool or be an interface to a third-party solver.

We've used CasADi’s capabilities in mathematical formulations of NLPs as an essential concept before formulating OCPs. CasADi formulates an NLP as follows:

\begin{equation}
     \mathop{minimize}_{x, p} \qquad  f(x, p)
    \label{eq:casadi_nlp}
\end{equation}

subject to:

\begin{equation}
   \mathop{x}_{-} \leq x \leq  \stackrel{-}{x}
    \label{eq:casadi_nlp_cons_1}
\end{equation}

\begin{equation}
    p =  \stackrel{-}{p}
    \label{eq:casadi_nlp_cons_2}
\end{equation}

\begin{equation}
     \mathop{g}_{-} \leq g(x, p) \leq \stackrel{-}{g}
    \label{eq:casadi_nlp_cons_3}
\end{equation}

where $f(x, p)$ and $g(x, p)$ denote the objective function and constraints function. Both functions depend on the decision variable $x$ and a known parameter p. The above-mentioned solution returns a primal $[x, p]$ and a dual $[x,g,p]$ solution, where the Lagrange multipliers are selected to be consistent with the Lagrangian function definition:

\begin{equation*}
    L(x, p, \lambda_x, \lambda_p, \lambda_g) :=
    \label{eq:Lagrangian}
\end{equation*}

\begin{equation}
    f(x, p) + \lambda_x^Tx + \lambda_p^Tp + \lambda_g^Tg(x, p)
    \label{eq:Lagrangian}
\end{equation}

According to the following assertion, this formulation drops all terms that do not depend on x or p and utilizes the same multipliers (but with different signs) for the inequality restrictions.




A positive multiplier indicates that the corresponding upper bound is active in this NLP formulation, and vice versa. Furthermore, p denotes the objective's parametric sensitivity to the parameter vector p. There are several numerical solvers that we can use to solve an NLP problem:

\begin{itemize}
    \item IPOPT \cite{biegler2009large}
    \item WORHP \cite{kuhlmann2018worhp} 
    \item SNOPT \cite{gill2005snopt}
\end{itemize}

In this article, we use an IPOPT solver (Nonlinear interior point method)  which is reported to be suitable for large-scale and sparse NLPs.

\section{Algorithmic Implementation}
The main idea in this section is to use the extracted kinematic model of the tractor-trailer system to design an NMPC local planner that is capable of handling both forward and backup maneuvering in the presence of static obstacles while respecting physical constraints. In the first subsection, we will be defining the concept of constraints on both systems states and control actions. Static obstacles are also considered inequality constraints. Later in the second subsection, we will be designing an NMPC local planner to handle path planning in both forward and backward maneuvering. This subsection will be followed by the third subsection in which we will be using the designed NMPC local planner to handle path following in both forward and backward maneuvering. Handling each task will be associated with multiple parking and maneuvering benchmark scenarios used in literature to compare our NMPC local planner to other state-of-the-art planners in the research community.

Based on the kinematic model of the standard tractor-trailer vehicle, the eulerian discretization method is used to obtain the discretized model of the system. Therefore the discretized model of the system will be in the following form:

\begin{equation}
    \begin{bmatrix}
    x_2(k+1)\\
    y_2(k+1)\\
    \theta_1(k+1)\\
    \theta_2(k+1)
    \end{bmatrix}
    =
    \begin{bmatrix}
    x_2(k)\\
    y_2(k)\\
    \theta_1(k)\\
    \theta_2(k)
    \end{bmatrix}
    +v
    \begin{bmatrix}
    cos(\theta_1)sin(\theta_2)\\
    -cos(\theta_1)cos(\theta_2)\\
    \frac{sin(\theta_1) + tan(\varphi)}{L_2}\\
    \frac{sin(\theta_1)}{L_2}
    \end{bmatrix}
    \label{eq:discretized_kinematic_model}
\end{equation}

Where $k$ represents the current iteration in the prediction horizon. $[x_2, y_2, \theta_1, \theta_2]^T$ is the state vector of the system. $v$ and $\varphi$ are the linear velocity at $S_2$ and the steering angle of the tractors’ front wheels, respectively. $L_2$ represents the trailer’s length and $N$ is the prediction horizon. Based on the discretized model, the nonlinear MPC local planner can be formulated as follows:

\begin{equation}
    minimize \, J(U)
    \label{eq:obj_func}
\end{equation}

subject to:

\begin{equation}
    \begin{matrix}
    x_{k+1} = f_d(x_{k-1}, u_k), & \forall k  \in [k, k+N] \\
    \delta U_{min} \leq \delta U \leq \delta U_{max}, & \forall k  \in [k, k+N-1] \\
    x_{min} \leq x \leq x_{max}, & \forall k  \in [k, k+N-1] \\
    U_{min} \leq U \leq U_{max}, & \forall k  \in [k, k+N-1]
    \end{matrix}
    \label{eq:obj_func_cons_1}
\end{equation}





Where $J$ is the objective function created over the prediction horizon. $u_k=[v, \phi]^T$ $\in$ $R^2$  is the control action vector at iteration $K$. $x_k$ is the state vector at iteration $k$ and $f_{d(x_{k-1}, u_k)}$ is the descretized model of the system. $U$ is the set of control actions over the prediction horizon: $U = [u_k, u_{k+1}, ..., u_{k+N-1}]$ $\in$ $R^{2^N}$.

There is also a constraint on the rate of the control actions. $\triangle U = [\triangle u_k, \triangle u_{k+1} ..., \triangle u_{k+N-1} ] \in R^{2^N}$ represents the change in the control actions over the prediction horizon. Therefore, at iteration $k$ , the associated $\triangle u$ is equal to $u_k - u_{k-1}$. 

To implement our non-linear MPC (NMPC) local planners, the CasADi optimization library is used. At each iteration, the objective function is created over the prediction horizon and is solved using the IPOPT method. CasADi provides both C++ and Python API for formulating optimization problems. We used Python to implement our non-linear MPC. The tractor steering system is based on the Ackermann model (Fig. X (TBD)) which is given in the following equation (TBD):

\begin{equation}
    cot(\varphi_o) = cot(\varphi_i) + \frac{w}{l}
    \label{eq:ackermann_equation}
\end{equation}

\begin{figure}[h!]
  \centering
  \includegraphics[scale=0.4]{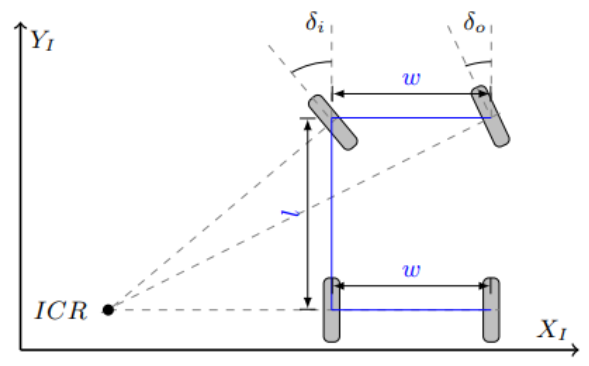}
  \caption{The Ackermann steering model \cite{daoud2019path}}
  \label{ackermann}
\end{figure}

Where the inner and outer steering angle is represented by $\phi_i$ and $\phi_o$, respectively. Figure \ref{ackermann} illustrates the dimensions of $l$ and $w$.

\subsection{Real-Time Local Planning Using NMPC}
In this section, we will consider forward and backward planning scenarios. During each scenario, the NMPC local planner receives the initial pose of $x_i = [x_i, y_i, \theta_{1,i}, \theta_{2,i}]$, target pose $x_o = [x_o, y_o, \theta_{1,o},\theta_{2,o}]$, and controller’s tuning parameters (e.g. weights). Once the objective function is solved over the prediction horizon, it outputs the sequence of control actions based on the minimum traveled distance between the initial and the target pose.

We’ve also defined a terminal condition for our NMPC local planner. Our terminal condition is based on $linalg.norm$ which is basically the distance between the current and target states of the system. Once the terminal condition is met, the controller considers the target as \textbf{met} and stops generating control actions.

We performed static obstacle avoidance missions in both forward and backward maneuvering. Therefore, we consider each obstacle to be a circular shape with a specific radius. In order to define hard constraints on the states of the system, we consider the tractor and the trailer’s body to be in a circular shape. Having said that, a marginal safety parameter $r_{safe}$ is also defined to prevent the vehicle from getting too close to the obstacle. 

\subsubsection{Static Obstacle Avoidance in Forward Local Planning}
\begin{figure}
    \centering
    \includegraphics[scale=0.46]{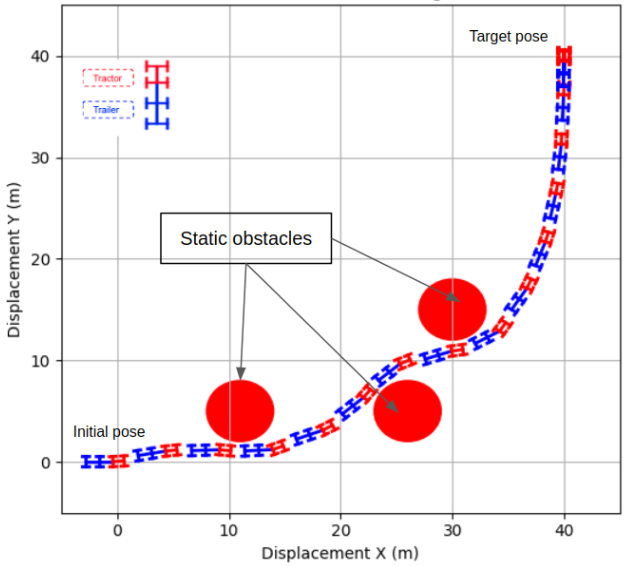}
    \caption{Results of forward maneuvering performed by NMPC local planner while avoiding multiple static obstacles}
    \label{fig:forward-obs-avd}
\end{figure}

\begin{figure}
    \centering
    \includegraphics[scale=0.28]{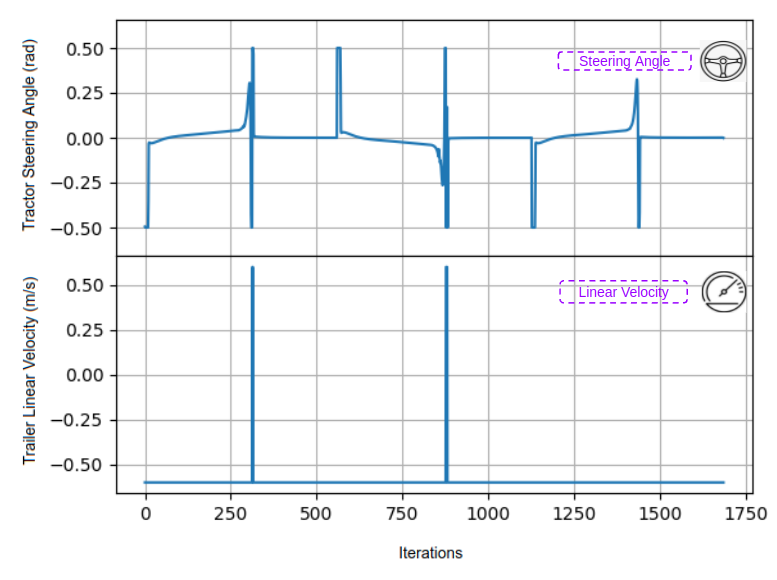}
    \caption{The first entry of control actions at each iteration over the prediction horizon during the forward planning scenario while avoiding static obstacles}
    \label{fig:forward-obs-avd-control-actions}
\end{figure}

In this subsection, the NMPC local planner is set to plan a series of control actions in order to drive the tractor-trailer system to a target point while avoiding multiple static obstacles in the operating environment. The system’s and planner’s parameters are listed in Table \ref{fig:forward-obs-avd}. Figure \ref{fig:forward-obs-avd} illustrates the local path generated by the planner. The obstacle diameters can be set to any desired value, individually. In this scenario, the planner was able to place the vehicle in its parking spot while the final error was reported to be at the minimum value of 0.05 m. Figure \ref{fig:forward-obs-avd-control-actions} illustrates the control actions generated by the planner at each iteration.

\begin{table}[h!]
\begin{center}
\resizebox{\linewidth}{!}{\begin{tabular}{||c c c c||} 
 \hline
 Parameter & Value &  Parameter & Value\\ [0.5ex] 
 \hline\hline
 Sampling Time & 0.2 & Max displacement X & [-5.0, 45.0] m \\ [0.5ex]
 \hline
 Prediction Horizon & 60 & Max displacement Y & [-5.0, 45.0] m\\ [0.5ex]
 \hline
 Tractor’s length & 1.9 & Max velocity & 0.2 m/s\\ [0.5ex]
 \hline
 Trailer’s length & 4.0 & Max Steering angle & 0.5 rad\\ [0.5ex]
 \hline
 Tractor-trailer’s widh & 1.0 & Max hitch angle & 0.7 rad\\ [0.5ex]
 \hline
 Obstacl’s diameter & 5.0 & Initial point & [0.0, 0.0, 0.0, 1.5707]\\ [0.5ex]
 \hline
 Target point & [40.0, 40.0, 0.0, 0.0] & Terminal error & 0.0.05 m\\ [0.5ex]
 \hline
\end{tabular}}
\caption{The system’s and planner’s parameters in forward planning while avoiding static obstacles in the operation environment}
\label{tab:forward-obs-avd}
\end{center}
\end{table}

\subsubsection{Path Following in Forward Local Planning}
Path following is the problem of following a series of discrete waypoints using a real-time local planner. Prior to starting the local planning process, these discrete waypoints are generated by a global planer to connect two or multiple targets in a static or dynamic environment. The high-level software of any autonomous driving system should be able to utilize GPS data or any kind of SLAM algorithm to localize the vehicle in the environment. Therefore, in this section, we assume that the vehicle pose is known and waypoints are already generated by a global planning algorithm. The system’s and planner’s parameters are listed in Table \ref{tab:forward-path-follow}.

\begin{table}[h!]
\begin{center}
\resizebox{\linewidth}{!}{\begin{tabular}{||c c c c||} 
 \hline
 Parameter & Value &  Parameter & Value\\ [0.5ex] 
 \hline\hline
 Sampling Time & 0.2 & Max displacement X & [-5.0, 25.0] m \\ [0.5ex]
 \hline
 Prediction Horizon & 60 & Max displacement Y & [-5.0, 20.0] m\\ [0.5ex]
 \hline
 Tractor’s length & 1.9 & Max velocity & 0.2 m/s\\ [0.5ex]
 \hline
 Trailer’s length & 4.0 & Max Steering angle & 0.5 rad\\ [0.5ex]
 \hline
 Tractor-trailer’s widh & 1.0 & Max hitch angle & 0.89 rad\\ [0.5ex]
 \hline
 Circle path diameter & 10.0 & Initial point & [5.0, 0.0, 0.0, 1.5707]\\ [0.5ex]
 \hline
 Target point & [5.0, 10.0, 0.0, -1.5707] & Terminal error & 0.1 m\\ [0.5ex]
 \hline
\end{tabular}}
\caption{The system’s and planner’s parameters in forward waypoint following scenario}
\label{tab:forward-path-follow}
\end{center}
\end{table}

\begin{figure}
    \centering
    \includegraphics[scale=0.35]{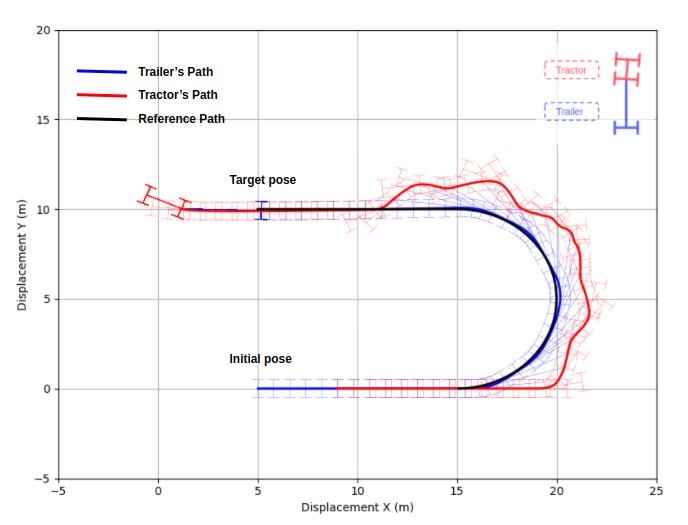}
    \caption{A sample result of forward waypoint following mission performed by our NMPC local planner. In this particular case, the circle diameter was 10.0 meters. The maximum positional error was reported to be 20 cm. Tractor and trailer paths are plotted in different colors.}
    \label{fig:forward-path-follow-xy}
\end{figure}

\begin{figure}
    \centering
    \includegraphics[scale=0.38]{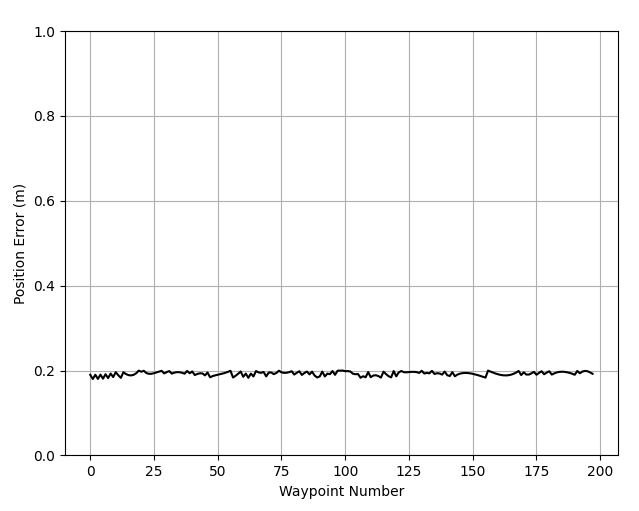}
    \caption{A sample of positional error of our proposed NMPC local planner in circular forward waypoint following mission. In this particular case, the circle diameter was 10.0 meters. The maximum positional error was reported to be 20.05 cm.}
    \label{fig:forward-path-follow-error}
\end{figure}

Figure \ref{fig:forward-path-follow-xy} illustrates the reference path along with the tractor and the trailer’s path as the result of circular waypoint following using our proposed NMPC local planner. We have tested waypoint-following ability of our proposed NMPC local planner on circular waypoints with different diameters. We consider the maximum positional error as a standard metric. The error metric combines both trailer’s position and orientation. Figure \ref{fig:forward-path-follow-error} illustrates the positional error tracking circular waypoints with different diameters. The density of the waypoints can be controlled in the source code. We considered the minimum of 2000 discrete points on the reference path.

As it can be seen from Figure \ref{fig:forward-path-follow-error}, the positional error is stable during the path-following mission. This behavior represents the stability of the local planner during the path-following mission. The positional error can be further reduced by tuning objective functions’ weights. On the other hand, if we relax the constraints on the states of the system, the RSME can be reduced to 0.12 m.

\subsubsection{Static Obstacle Avoidance in Backward Local Planning}
Backward maneuvering is considered to be a difficult task in local planning. Technically speaking, in order to prevent the jackknife situation, the planner needs to take the hitch angle between the tractor and the trailer’s body into account. In order for the planner to perform a stable and successful backward maneuvering mission, it needs to consider the constraints on the hitch angle very strictly. In our system formulation, we consider the hitch angle as a system state which allows us to put constraints on its value during the minimization process. We show the ability of our proposed local planner in performing static obstacle avoidance in backward maneuvering scenarios. The system and planner parameters were set to be the same as forward planning parameters mentioned in Table \ref{tab:forward-obs-avd} in the previous section. The only difference is the vehicle orientation at the initial and target positions.

\begin{figure}
    \centering
    \includegraphics[scale=0.3]{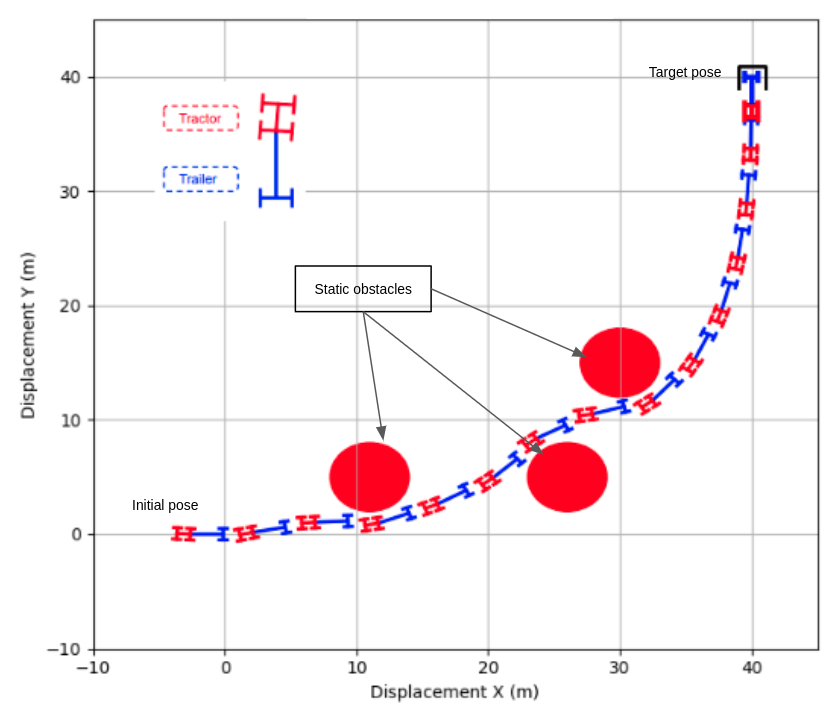}
    \caption{Results of backward maneuvering performed by NMPC local planner in presence of static obstacles. The planner in able to plan the shortest possible path while avoiding multiple static obstacles.}
     \label{fig:3-backward-obs-xy}
\end{figure}

\begin{figure}
    \centering
    \includegraphics[scale=0.3]{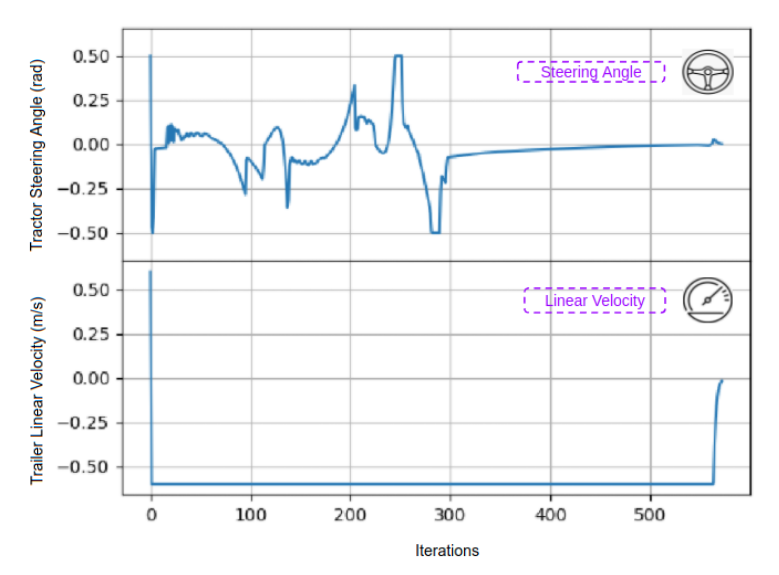}
    \caption{ The first entry of control actions at each iteration over the prediction horizon in backward planning while avoiding multiple static obstacles}
    \label{fig:backward-obs-control-actions}
\end{figure}

\begin{figure}
    \centering
    \includegraphics[scale=0.37]{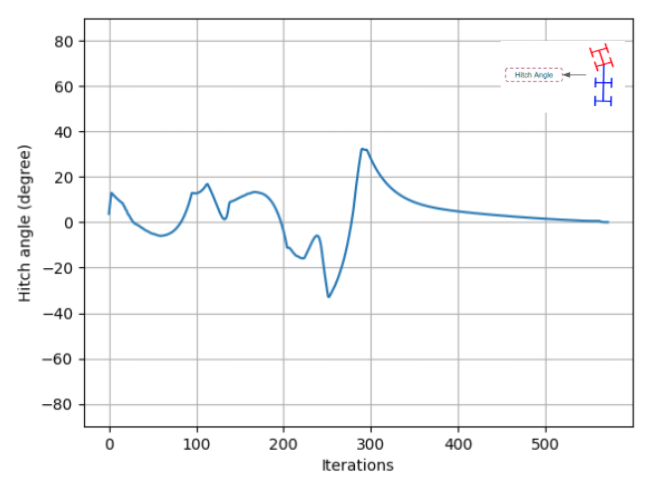}
    \caption{Hich angle of the tractor-trailer system at each iteration over the prediction horizon in backward planning while avoiding multiple static obstacles}
    \label{fig:backard-obs-hitch-angle}
\end{figure}

As it can be seen from Figure \ref{fig:3-backward-obs-xy}, the planner is able to navigate the vehicle between multiple obstacles and place it in the target position. The control actions and hitch angle are also illustrated in Figure \ref{fig:backward-obs-control-actions} and \ref{fig:backard-obs-hitch-angle}, respectively. In our proposed local planner, as mentioned before, the hitch angle is considered to be one of the system’s states which allows us to put constraints on its maximum and minimum value during the planning process. As a result, the vehicle is fully stable and will never experience a jackknife situation.

\subsubsection{Path Following in Backward Local Planning}
In this section, we present the ability of our NMPC local planner in the circular waypoint following the scenario in backup maneuvering. Table \ref{tab:backward-path-follow}, illustrates the planner parameters and vehicle dimensions. We also calculate the deviation of the vehicle’s position from the reference path in the shape of normal error and RSME over the entire mission.

\begin{table}[h!]
\begin{center}
\resizebox{\linewidth}{!}{\begin{tabular}{||c c c c||} 
 \hline
 Parameter & Value &  Parameter & Value\\ [0.5ex] 
 \hline\hline
 Sampling Time & 0.2 & Max displacement X & [-5.0, 25.0] m \\ [0.5ex]
 \hline
 Prediction Horizon & 60 & Max displacement Y & [-5.0, 20.0] m\\ [0.5ex]
 \hline
 Tractor’s length & 1.9 & Max velocity & 0.2 m/s\\ [0.5ex]
 \hline
 Trailer’s length & 4.0 & Max Steering angle & 0.5 rad\\ [0.5ex]
 \hline
 Tractor-trailer’s widh & 1.0 & Max hitch angle & 0.89 rad\\ [0.5ex]
 \hline
 Circle path diameter & 10.0 & Initial point & [5.0, 10.0, 0.0, 1.5707]\\ [0.5ex]
 \hline
 Target point & [5.0, 0.0, 0.0, -1.5707] & Terminal error & 0.1 m\\ [0.5ex]
 \hline
\end{tabular}}
\caption{The system’s and planner’s parameters in backward waypoint following scenario}
\label{tab:backward-path-follow}
\end{center}
\end{table}

\begin{figure}
    \centering
    \includegraphics[scale=0.34]{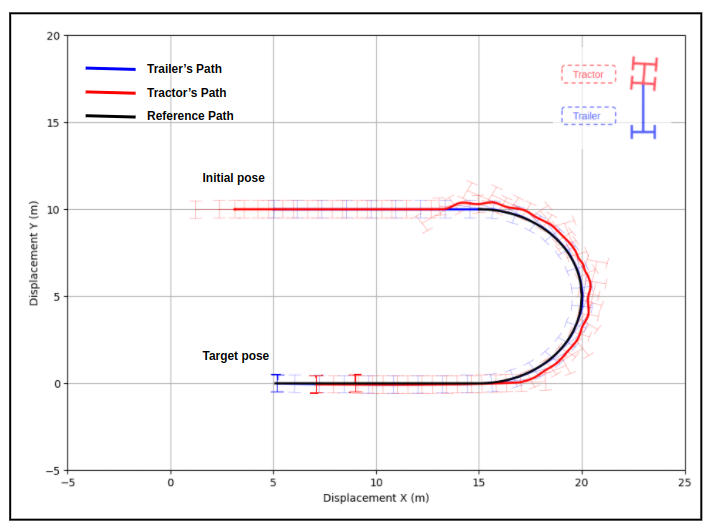}
    \caption{A sample result of backward waypoint following mission performed by our NMPC local planner in back up maneuvering. In this particular case, the circle diameter was 10.0 meters. The maximum positional error was reported to be 11.1 cm. Tractor and trailer paths are plotted in different colors.}
    \label{fig:backward-path-follow-xy}
\end{figure}

\begin{figure}
    \centering
    \includegraphics[scale=0.4]{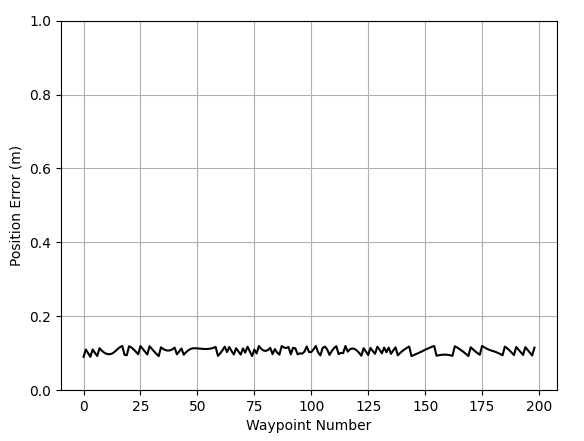}
    \caption{A sample of positional error of our proposed NMPC local planner in backward circular waypoint following mission. In this particular case, the circle diameter was 10.0 meters. The maximum positional error was reported to be 11.1 cm.}
    \label{fig:backward-path-follow-error}
\end{figure}

Figure \ref{fig:backward-path-follow-xy} and \ref{fig:backward-path-follow-error} illustrate the tractor-trailer path and the positional error over the entire mission while performing reference waypoints following in backward maneuvering. Our proposed NMPC local planner was able to perform the mission with maximum positional error of 11.1 cm. The RSME over the entire mission was reported to be 10.6 cm.

\section{Mechanistic Modeling and Simulation}
\begin{figure}
    \centering
    \includegraphics[scale=0.28]{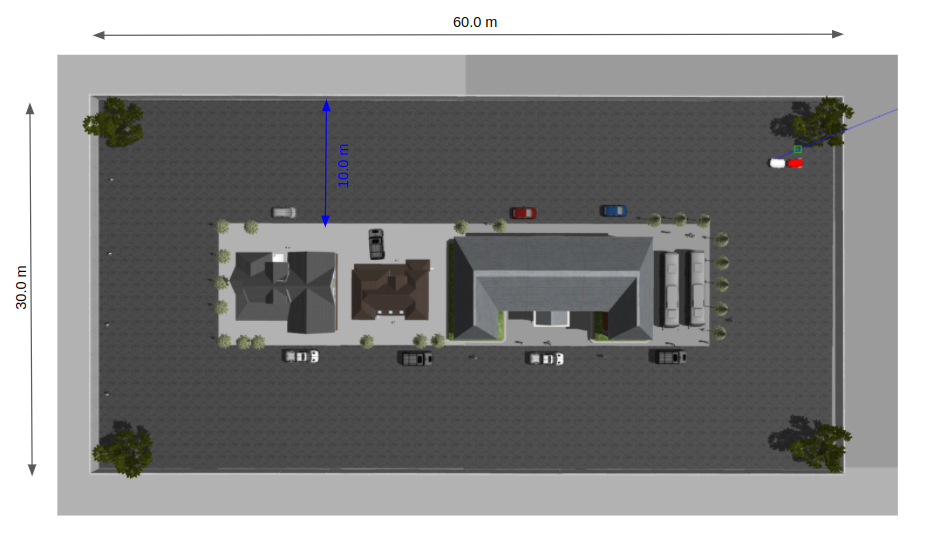}
    \caption{The simulated urban environment in Gazebo}
    \label{fig:gazebo_robo_city}
\end{figure}

\begin{figure}
    \centering
    \includegraphics[scale=0.35]{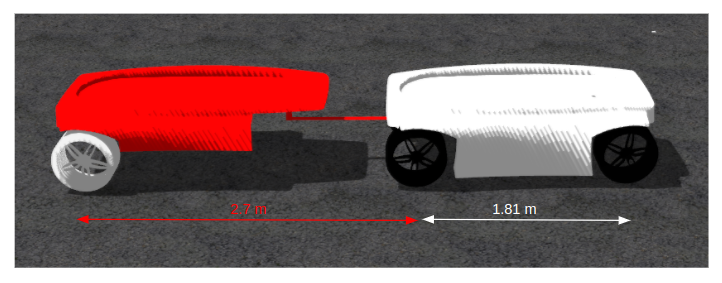}
    \caption{The simulated tractor-trailer vehicle in Gazebo}
    \label{fig:gazebo_vehicle}
\end{figure}

Simulators have been crucial in robotics research for the rapid and efficient testing of new concepts, methods, and algorithms. A robotics simulator is used to design embedded applications for a robot without physically relying on the actual machine. These applications can, in certain cases, be transferred to the real robot without modification\cite{castillo2010introductory}. Gazebo, as a real-time robotic simulator with the physics engine, is integrated with ROS enabling researchers to implement their algorithms in the simulation before running it on a real robot. There are libraries for physics simulation, graphics, user interface, communication, and sensor configuration in the gazebo simulator. Gazebo can simulate a robot in a three-dimensional environment. It creates both accurate sensor feedback and physically plausible object interactions (it includes an accurate simulation of rigid-body physics). Code written to drive a physical robot can be executed on an artificial equivalent by realistically replicating robots and settings. Our final goal in this section is to integrate our NMPC local planner with Autoware.ai (A world-class autonomous driving platform under ROS). Autoware.ai is created to be a comprehensive software stack for autonomous vehicles using open-source software features. The self-driving algorithmic implementation mentioned here is intended for vehicles operated in cities rather than on freeways or highways. Autoware.ai does not support tractor-trailer vehicles in its software stack. Specifically, in terms of motion planning algorithms, only single vehicles are supported. Our main effort in this section was adding an NMPC local planner to Autoware's motion planning algorithms to handle local planning missions of tractor-trailer vehicles. Therefore, we designed a complete simulation environment in Gazebo. Afterward, we integrated our simulated vehicle with Autoware's vehicle model, giving us the ability to customize other parts of the software for our specific application. The simulated urban environment and tractor-trailer vehicle are illustrated in Figures \ref{fig:gazebo_robo_city} and \ref{fig:gazebo_vehicle}, respectively. 

The vehicles sensor configuration can also be defined using URDF format. A Velodyne VLP-16 3D laser scanner and an RGB camera are mounted on the vehicle. The laser scanner is mounted in the X-Y position on the king pine and its height from the ground is 1.8 m. The camera is mounted on the front part of the tractor's body. 
The simulated tractor-trailer vehicle is operated by two control signals of rotational velocity of the tractor's rear wheels and the steering angle of the tractor's front wheels. Therefore, we have defined PID controller on their related joints. We have used the \textbf{controller-manager} package of ROS to deploy the controllers on the related joints. That gives us the ability to send control action signals from our NMPC local planner algorithm directly to the vehicle in the simulation environment.

\subsection{Mapping and Localization}
\begin{figure}
    \centering
    \includegraphics[scale=0.15]{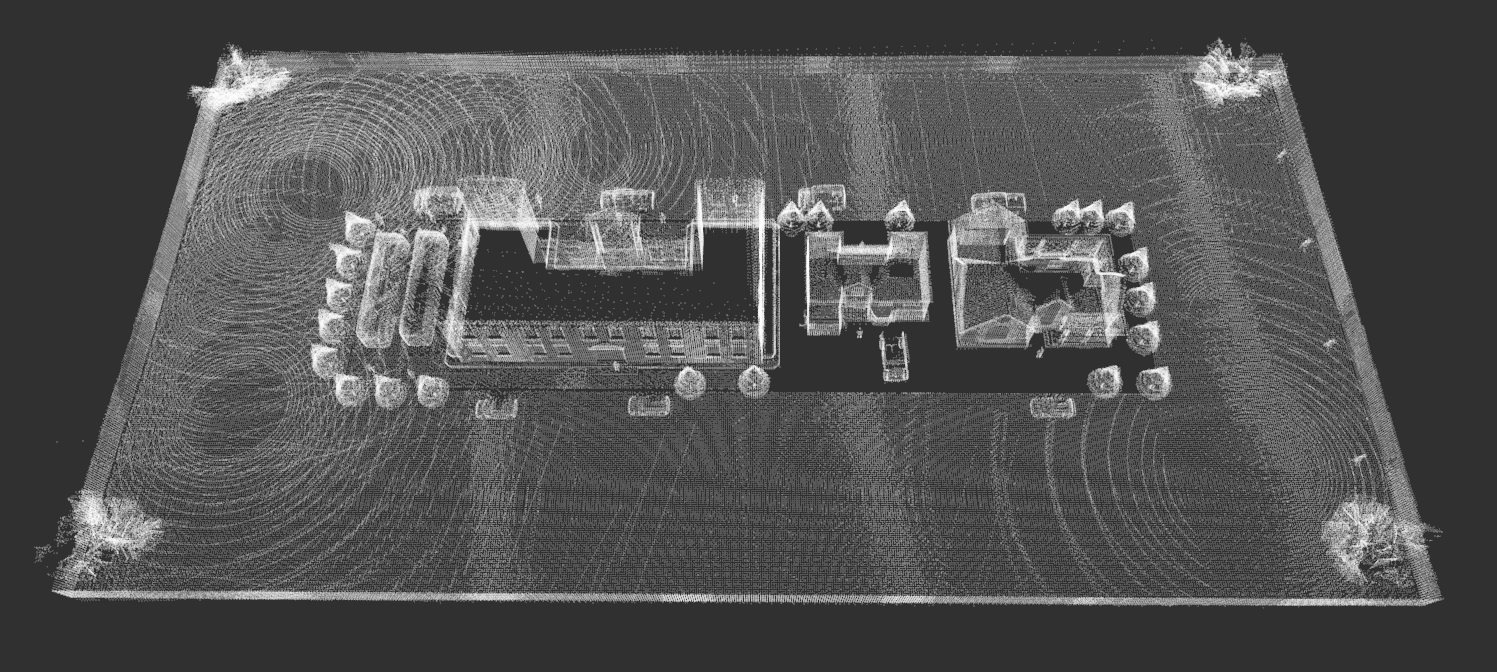}
    \caption{The generated 3D point-cloud map of the simulated urban environment. Autoware's ros-package implementation of NDT mapping is used for this purpose}
    \label{fig:rviz-map}
\end{figure}

Autoware is basically using 3D LiDAR scanners to recognize the road and surrounding environment. LiDAR scanners calculate distances to objects by lighting a target with pulsed lasers and timing the reflected pulses. LiDAR scanner point cloud data can be used to produce digital 3D representations of scanned objects. Cameras are mostly employed to recognize traffic lights and extract additional information from scanned items. Processing point-cloud data is computationally expensive. Therefore, Autoware filters and pre-processes the raw point cloud data obtained by LiDAR scanners to achieve real-time processing. To create the 3D point-cloud map of the environment, localize the vehicle, and detect objects in a point cloud, voxel grid filtering is implemented in a specific package. This type of filtering reduces the size of the point cloud by replacing a group of points in a cubic lattice grid with their centroid.
For localization, Autoware mostly employs the Normal Distributions Transform (NDT) \cite{biber2003normal} algorithm. Because the computing cost of the NDT algorithm is not driven by map size, high-definition and high-resolution 3D maps can be used at wider scales. NDT algorithm \cite{biber2003normal} can be used to calculate the relative position between two point clouds using this downsampled point cloud. This result is used by Autoware to obtain an exact position between the 3D map and the LiDAR scanner mounted on the car. In this article, Autoware's \textbf{ndt-mapping} package is utilized to generate the 3D point-cloud map of the simulated urban environment. Figure \ref{fig:rviz-map} illustrates the generated 3D map. Instead of the entire generated 3D map, the matching procedure is performed against the prior LiDAR scan to build a map using this technique. Once the transformation is obtained, the current scan is transformed to the prior scan's coordinate system, and the point cloud is summed. This is done repeatedly when constructing the map.

\begin{figure}
    \centering
    \includegraphics[scale=0.25]{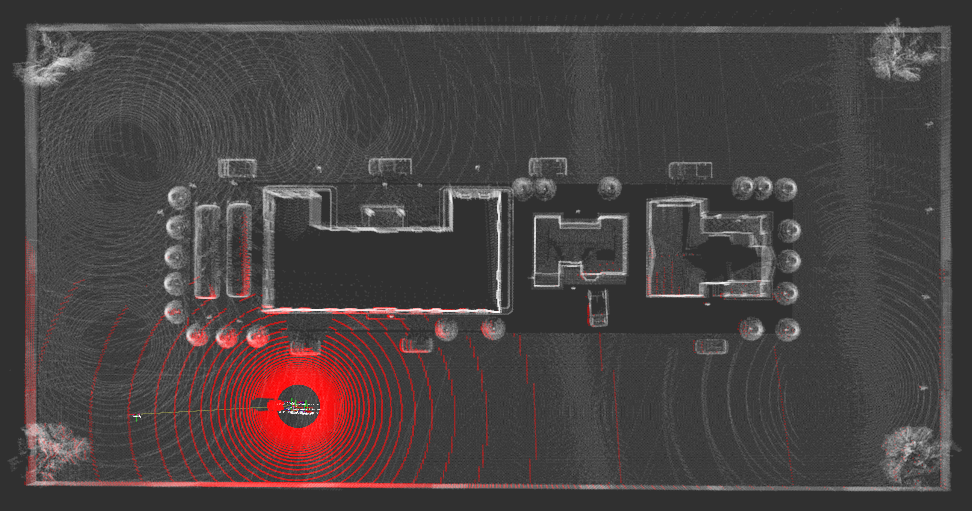}
    \caption{The localized vehicle in the 3D point-cloud map of the urban environment}
    \label{fig:rviz-localization}
\end{figure}

In order to localize the vehicle in the map, the Autoware's \textbf{ndt-matching} package was used to extract the accurate position of the vehicle in the map. The typical inaccuracy is a few centimeters. This pose data will be used by our NMPC local planner package. Figure \ref{fig:rviz-localization} illustrates the vehicle while it is localized in the 3D point-cloud map.

\subsection{Perception}
In Autoware, perception is mostly achieved utilizing point cloud data received from LiDAR scanners. The nearest neighbors approach is used to pre-process and segment the point cloud. This procedure computes the Euclidean distance between points in 3D space based on a distance threshold. After clustering the point cloud, the distance between the surrounding objects and the ego-vehicle can be estimated. Furthermore, using the distance between each cluster and the categorized 2D image processing techniques, objects can be monitored over time to increase perceptual information.

\begin{figure}
    \centering
    \includegraphics[scale=0.4]{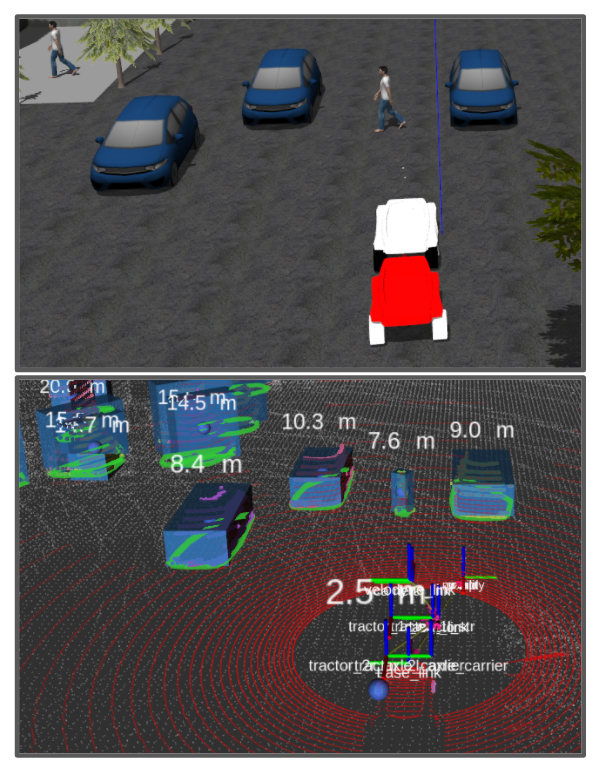}
    \caption{Top: Our simulated urban environment, Bottom: Extracted on-road obstacles in our simulated urban environment using Autoware's \textbf{euclidean-cluster-detection} package.}
    \label{fig:eu-cluster-detection}
\end{figure}

In this article, we used Autoware's \textbf{euclidean-cluster-detection} package to detect on-road obstacles. This package receives the down-sampled LiDAR's point cloud and outputs the segmented clusters. As it can be seen
from Figure \ref{fig:eu-cluster-detection}, Autoware's \textbf{euclidean-cluster-detection} package is capable of detecting pedestrians, cars, trees, and buildings of different sizes in real-time. Later, the extracted on-road obstacles will be fed to our NMPC local planner in obstacle avoidance missions.

\subsection{Real-Time Local Planning Using the Proposed NMPC}
As discussed before, our main contribution in this section is to add a tractor-trailer vehicle to Autoware's autonomous driving software stack \cite{Autoware2013}. Therefore, we have designed a full simulated urban environment, a standard tractor-trailer vehicle, and most importantly, an NMPC local planner to handle the motion planning task for this type of autonomous vehicle. 

\begin{figure}
    \centering
    \includegraphics[scale=0.185]{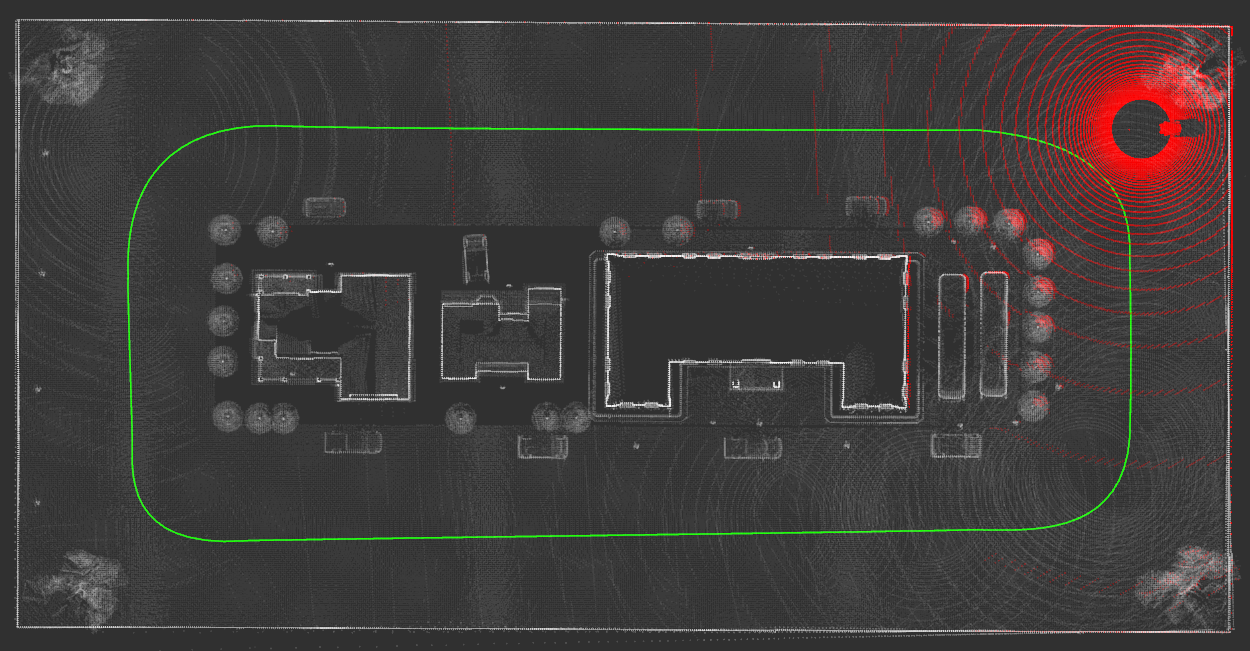}
    \caption{A simple Vector Map (e.g. A closed curve projected in the yellow color) used by our NMPC local planner to perform path following missions in both forward and backward maneuvering}
    \label{fig:vector-map}
\end{figure}

The planning part of Autoware's software stack can be divided into two main branches of \textbf{Mission Planning} and \textbf{Local Planning}. The mission planner uses a rule-based system to predict trajectories based on driving conditions such as lane changes, merges, and passing. Mission planning also includes navigation from the current location to the destination. The high-definition 3D map data includes static road characteristics that can be used for mission planning. Once the global path is established, the motion planning module is activated to create a plan for local trajectories. The local planner's primary task is to follow the center lines of the lanes over the route created by the map navigation system, which uses high-definition 3D map information. In our application, we have designed our global path using Vector Map standard \cite{lee2013vector}. Figure \ref{fig:vector-map} illustrates a simple Vector Map that will be used by our NMPC local planner to perform path-tracking missions in both forward and backward maneuvering. 

The local planner is responsible for developing local viable trajectories in addition to the specified global trajectories, taking into account the vehicle states, drivable zones indicated by the 3D map, surrounding objects, traffic restrictions, and the desired target. Autoware uses \textbf{Pure-pursuit} algorithm in order to perform local planning (e.g. generating the control actions in order to follow a trajectory/path. Also, tractor-trailer systems are not supported by this algorithm. More importantly, \textbf{Pure-pursuit} does not take the on-road obstacles into account. It is simply designed to receive a reference path and outputs control actions to follow the path.

\subsubsection{Path Following in Forward and Backward Maneuvering}
\begin{figure}
    \centering
    \includegraphics[scale=0.13]{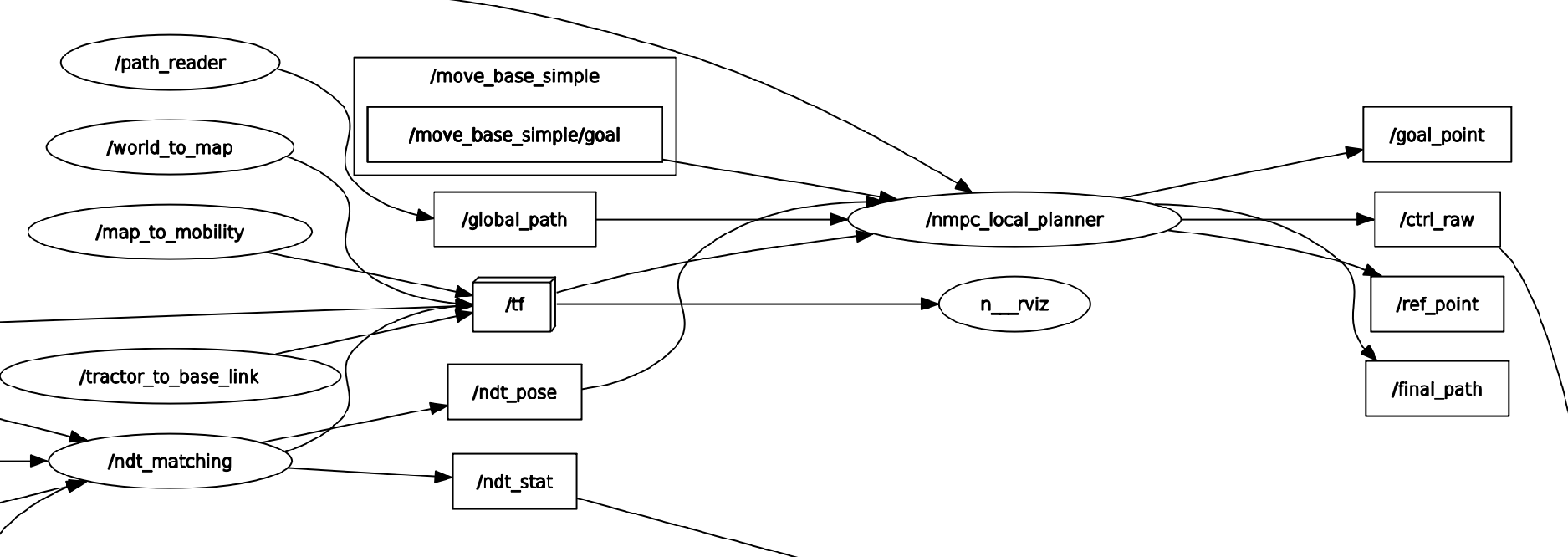}
    \caption{The RQT graph that indicated the input and output data to our NMPC local planner. In this scenario, the planner is assigned to follow a reference path (/global-path) in both forward and backward maneuvering. Therefore, the environment is considered to be obstacle-free.}
    \label{fig:nmpc_rqt_graph}
\end{figure}

In this section, we will be presenting our NMPC local planner's results in handling path-following missions using tractor-trailer vehicles. Figure \ref{fig:nmpc_rqt_graph} illustrates the data flow of the NMPC local planner node in ROS. The NMPC local planner receives the global path, frame transformations(/tf), and the real-time position of the vehicle (/ndt-pose) in the map. The most important output signals are the control actions (/ctrl-raw). These command signals are sent directly to the Vehicle actuators in the gazebo simulator. Algorithm \ref{alg:path-following} explains our path-following mission. First, it receives NDT-Pose, Vecor-Map, target pose $x_s$, and look-ahead distance at each iteration in real-time. Then it extracts the nearest waypoint $NP$ from the vector map $VM$. Based on the value of $NP$ and $\lambda$, it extracts the next target pose $x_s$ on the vctor map $VM$. Finally, $x_s$ will be sent to the NMPC controller (Algorithm \ref{alg:nmpc-controller}) to return the control actions. In our simulated environment, the maximum tracking error during this mission was reported to be 46.3cm. The minimum real-time execution factor of the simulator was reported to be 0.83.

\begin{algorithm}
\caption{Path Following}\label{alg:path-following}
\begin{algorithmic}
\State \textbf{Input. 1:} NDT-Pose (Real-time position of the vehicle in the map)
\State \textbf{Input. 2:} Vector Map
\State \textbf{Input. 3:} Target Pose
\State \textbf{Input. 4:} $L: $ Look-ahead distance

\State \textbf{Output:} Control Actions (linear velocity and steering angle)
\For{\texttt{each iteration over the prediction horizon}}
    \State $Error = norm(x_s - x_k)$
    \If{$Error > Min Error$}
        \State - Calculate Labmda: $\lambda = L - (L/(1 + Error))$
        \State - Extract the nearest waypoint to the 
        \State \, vehicle: $NP = VM^1(x_k)$
        \State - Extract the target pose: $xs = VM(NP + \lambda)$
        \State - Execute Algorithm \ref{alg:nmpc-controller} and calculate
        \State \, the control actions
    \EndIf
\EndFor

\end{algorithmic}
\end{algorithm}

\begin{figure}
    \centering
    \includegraphics[scale=0.4]{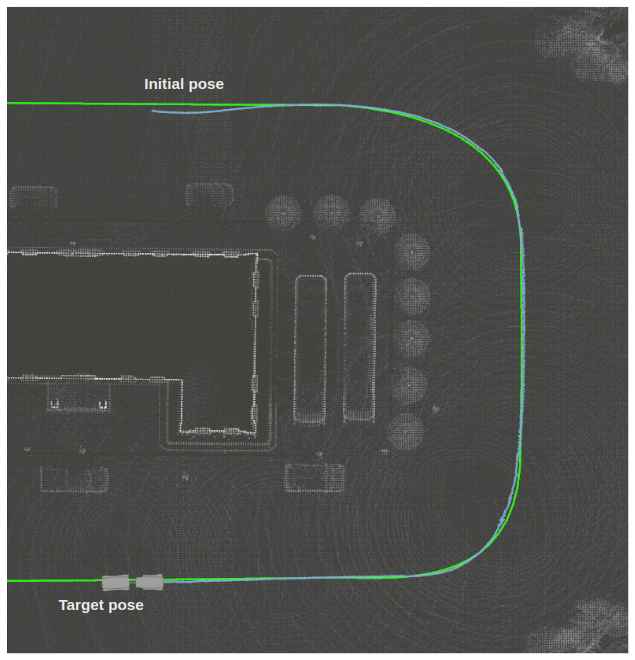}
    \caption{Path following mission in forward maneuvering. The results are shown in RViz which is a ROS-BAsed visualization tool. The yellow and purple lines indicate the global path and the vehicle's path, respectively.}
    \label{fig:rviz_forward_path_following}
\end{figure}

\subsubsection{Static Obstacle Avoidance in Forward and Backward Maneuvering}
\begin{figure}
    \centering
    \includegraphics[scale=0.19]{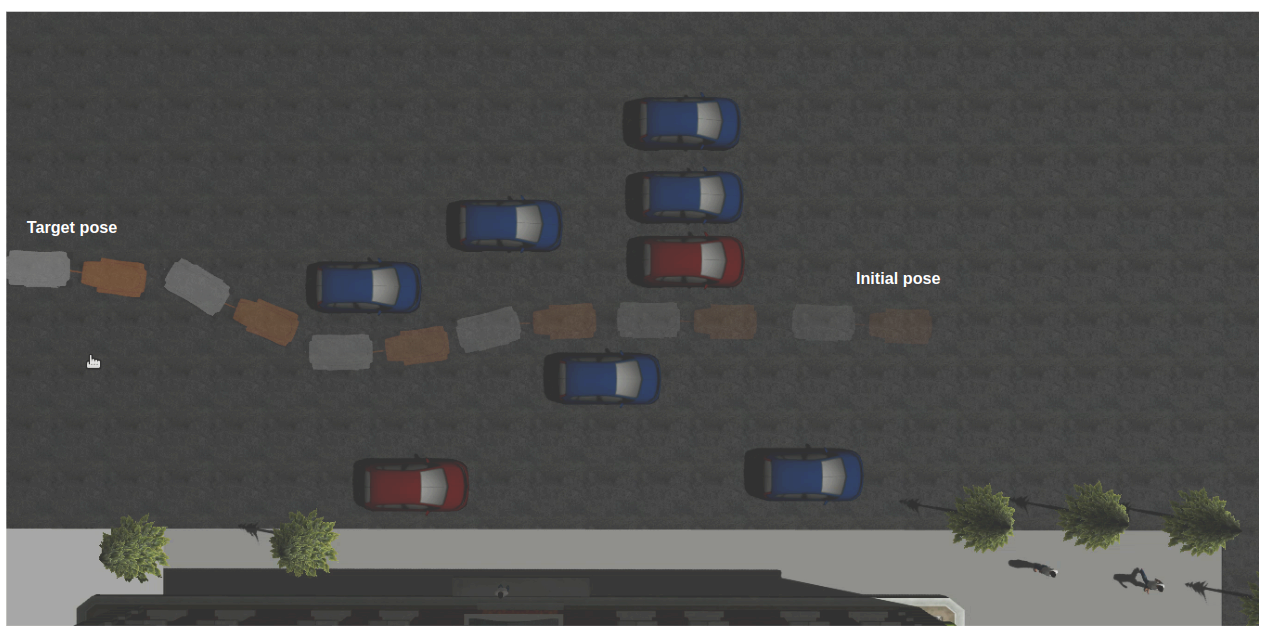}
    \caption{Multiple static obstacle avoidance mission in forward maneuvering}
    \label{fig:Forward_obstacle_avoidance_gazebo}
\end{figure}

\begin{figure}
    \centering
    \includegraphics[scale=0.22]{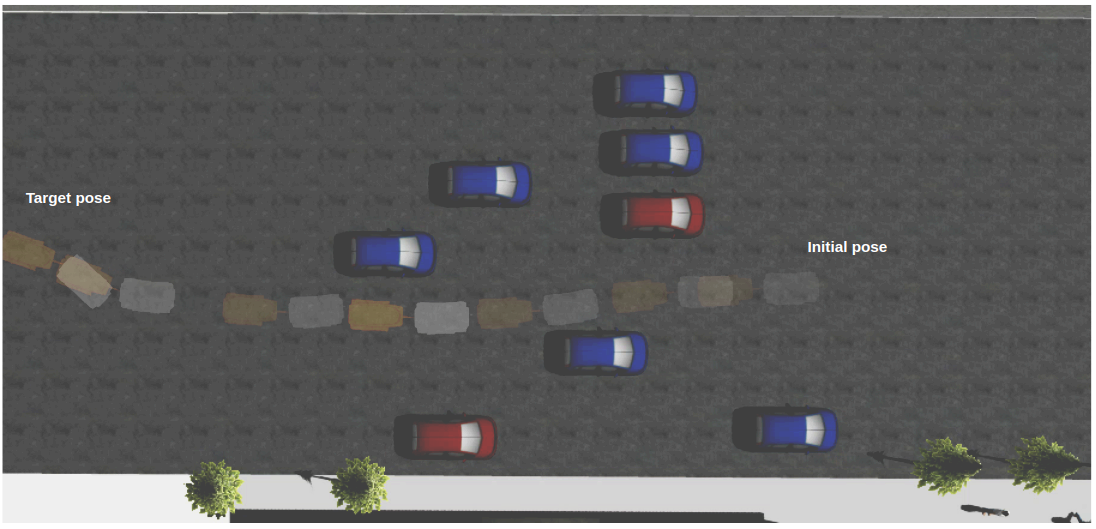}
    \caption{Multiple static obstacle avoidance mission in backward maneuvering}
    \label{fig:Backward_obstacle_avoidance_gazebo}
\end{figure}

In this section, we will be presenting the results of multiple obstacle avoidance missions performed by our NMPC local planner using our simulated tractor-trailer vehicles. We assigned the planner to perform the obstacle avoidance mission in both forward and backward maneuvering. Algorithm \ref{alg:obstacle-avoidance} explains the obstacle avoidance mission. After receiving the vehicle's position in the map, the real-time position of on-road obstacles are extracted using Algorithm \ref{alg:euclidean_clustering_process}. An example of on-road obstacle extraction is illustrated in Figure \ref{fig:eu-cluster-detection}. At the next step, the local path is extracted and transformed to the vector map. Finally, Algorithm \ref{alg:path-following} is executed to generate the control actions. Figures \ref{fig:Forward_obstacle_avoidance_gazebo} and \ref{fig:Backward_obstacle_avoidance_gazebo} illustrate the results of executing these missions.

\begin{algorithm}
\caption{Obstacle Avoidance}\label{alg:obstacle-avoidance}
\begin{algorithmic}
\State \textbf{Input. 1:} NDT-Pose (Real-time position of the vehicle in the map)
\State \textbf{Input. 2:} On-road obstacle positions in real-time
\State \textbf{Input. 3:} Target Pose $x_s$

\State \textbf{Output:} Control Actions (linear velocity and steering angle)
\For{\texttt{each iteration} $n \in N$}
    \State - Read the current position $x_k$ of the vehicle 
    \State - Execute  to get the obstacle positions
    \State - Calculate the Error: $Error = norm(x_s - x_k)$
    \If{$Error > MinError$}
        \State - Extract the local path and control actions
        \State \, using NMPC local planner
        \State - Transform the local path into Vector Map $VM$
        \State - Apply the control actions
    \EndIf
\EndFor

\end{algorithmic}
\end{algorithm}

\section{Summary of Objectives and Contributions}
\subsection{Objective 1: NMPC Local Planner for Tractor-Trailer Vehicles in Forward and Backward Maneuvering}
Prior to implementing our algorithm in a simulated environment, we decided to design and test our NMPC local planner in a headless python simulation. In the headless simulation, the current position of the vehicle was assumed to be its last position after executing the last control actions. Having said that, we evaluated the planner on a real-time basis. We considered both forward and backward planning missions in path following and static obstacle avoidance scenarios. Throughout the entire research, we've tried to test our planner against benchmark scenarios \cite{liu2014multi}. 

The first objective of this section was to adapt a single kinematic model of a tractor-trailer vehicle to perform both forward and backward maneuvering. Therefore, our planner was capable of making an initial decision about the direction of every planning mission (e.g. forward or backward). Real-time online optimization was our second objective in this section. In every non-linear optimal control problem, there is a large-scale objective function to be solved at each iteration over the prediction horizon. Depending on the kinematic/dynamic complexity of the system, solving this objective function can be time-consuming. Therefore, the solution is always associated with delay and latency. This problem is considered to be a research bottleneck in optimal control theory to handle real-time systems such as autonomous driving platforms \cite{qian2014manipulation}. Considering a non-linear kinematic model of the standard tractor-trailer vehicle, the objective function can be highly non-linear over the prediction horizon at each sampling time. That makes the minimization process extremely difficult for the solver \cite{mohamed2018optimal}. Our approach to pass this bottleneck was to transform the optimal control problem (OPT) into a non-linear programming problem (NLP). This transformation was performed using both single \cite{andersson2019casadi} and multiple shooting approaches \cite{bock1984multiple}. A single shooting approach is capable of employing fully adaptive, error-controlled state-of-the-art ODE or DAE solvers. It has few optimization degrees of freedom even for large ODE or DAE systems, and the fact that it only requires initial assumptions for control degrees of freedom. The flaws are that we can't use knowledge of the state trajectory $x$ in the initialization (for example, in path following and trajectory tracking), and the ODE solution $x(t; q)$ can rely very non-linearly on $q$. Therefore, we found the Single Shooting approach not suitable to deal with unstable systems such as tractor-trailer vehicles in backward maneuvering.

On the other hand, the Multiple Shooting approaches attempt to combine the features of a simultaneous method, such as collocation, with the major feature of single shooting, notably the ability to apply adaptive, error-controlled ODE solvers. In contrast to collocation, Multiple Shooting can use adaptive ODE/DAE solvers to combine adaptivity with fixed NLP dimensions. The ODE solution is frequently the most computationally expensive portion of each Sequential Quadratic Programming (SQP) loop, yet it is also the easiest to parallelize. NLP has a smaller dimension than collocation but is less sparse. The expense of the underlying ODE solution, along with the loss of sparsity, results in theoretically larger costs per SQP iteration than in collocation. Multiple Shooting, on the other hand, is a strong rival to direct collocation in terms of CPU time per iteration due to the ability to leverage efficient state-of-the-art ODE/DAE solvers and their innate adaptivity. 

We transformed our OPT to an NLP using both Single and Multiple Shooting approaches. In the forward obstacle avoidance mission (Figure \ref{fig:forward-obs-avd}), both approaches were able to perform the planning missions successfully. In terms of running time, Multiple Shooting was running faster. In the backward obstacle avoidance mission, the vehicle can easily jackknife. Therefore, putting a hard constraint on the hitch angle made it difficult for the solver to find an optimal solution using the Single Shooting approach. We found the Multiple Shooting approach stronger in stabilizing the system and faster in running time.

Our next scenario was to perform the path-following mission in both forward and backward maneuvering. We used a benchmark path offered by \cite{wu2017path} and \cite{prasad2016geometric} to compare the performance of our NMPC local planner. Utilizing the Single Shooting, the maximum tracking error was reported to be 25.2cm and 20.1cm in forward and backward path planning scenarios, respectively. However, the results showed a better path planning performance in using the Multiple Shooting approach. The maximum tracking error was reported to be 16.4cm and 11.1cm in forward and backward maneuvering, respectively. We observed that using the Multiple Shooting approach, our NMPC local planner was able to perform a backward path-following mission with the same tracking error that was previously reported by \cite{wu2017path}. Having said that they've not implemented their planner in a real-time simulation.

\subsection{Objective 2: A Fully-Automated Simulation Environment to Deploy Tractor-Trailer Vehicles}
To the best of our knowledge, we were not able to find real-time simulated tractor-trailer vehicles and their associated local planners. Most of the research works were conducted using a python/C++ headless simulated environment. Therefore, their result was not validated using state-of-the-art simulators in ROS. Our objective in this section was to design a real-time simulated environment (Gazebo-ROS) to evaluate our NMPC local planner in real-time. Considering the physics engine of the Gazebo simulator, the tractor-trailer vehicle was designed with the same dynamic and kinematic specifications as a real-world vehicle. In terms of sensor configuration, the vehicle was equipped with Velodyne VLP-16 3D laser scanners, IMU, Camera, and Radar plugins. therefore, we were able to use NDT mapping and NDT matching to create point-cloud maps and localize the vehicle in the map. We also utilized the Euclidean Cluster Detection algorithm to detect on-road obstacles. Therefore, our NMPC local planner was fed by the real-time feedback of the position of the vehicle on the map and on-road obstacles around the vehicle. We went one step further and integrated our simulated environment and NMPC local planner with a world-class autonomous driving software stack called Autoware.ai. 
Our NMPC local planner was able to perform both forward and backward path following and obstacle avoidance missions in the simulated environment.

\subsection{Future Works}
During our research and investigation, we observed that our planner is not capable of recovering from the jackknife situation. For instance, in case the vehicle is in a jackknife situation in backward maneuvering, our NMPC planner is not able to recover from the failuFig. 21. Multiple static obstacle avoidance mission in backward maneuvering.
In terms of practical implementation, our next plan is to deploy our local planner on a real autonomous tractor-trailer vehicle. Having said that, our simulated environment is highly identical to the real-world vehicle. Therefore, the most challenging part will be designing a low-level control system for the vehicle to control actuators and to receive high-level control signals from the planner.

\bibliography{main.bib}

\begin{thebibliography}{10}
\providecommand{\url}[1]{#1}
\csname url@samestyle\endcsname
\providecommand{\newblock}{\relax}
\providecommand{\bibinfo}[2]{#2}
\providecommand{\BIBentrySTDinterwordspacing}{\spaceskip=0pt\relax}
\providecommand{\BIBentryALTinterwordstretchfactor}{4}
\providecommand{\BIBentryALTinterwordspacing}{\spaceskip=\fontdimen2\font plus
\BIBentryALTinterwordstretchfactor\fontdimen3\font minus
  \fontdimen4\font\relax}
\providecommand{\BIBforeignlanguage}[2]{{%
\expandafter\ifx\csname l@#1\endcsname\relax
\typeout{** WARNING: IEEEtran.bst: No hyphenation pattern has been}%
\typeout{** loaded for the language `#1'. Using the pattern for}%
\typeout{** the default language instead.}%
\else
\language=\csname l@#1\endcsname
\fi
#2}}
\providecommand{\BIBdecl}{\relax}
\BIBdecl

\bibitem{worthmann2015model}
K.~Worthmann, M.~W. Mehrez, M.~Zanon, G.~K. Mann, R.~G. Gosine, and M.~Diehl,
  ``Model predictive control of nonholonomic mobile robots without stabilizing
  constraints and costs,'' \emph{IEEE transactions on control systems
  technology}, vol.~24, no.~4, pp. 1394--1406, 2015.

\bibitem{lu2022real}
H.~Lu, Q.~Zong, S.~Lai, B.~Tian, and L.~Xie, ``Real-time perception-limited
  motion planning using sampling-based mpc,'' \emph{IEEE Transactions on
  Industrial Electronics}, 2022.

\bibitem{bin2012constrained}
Y.~Bin and T.~Shim, ``Constrained model predictive control for backing-up
  tractor-trailer system,'' in \emph{Proceedings of the 10th World Congress on
  Intelligent Control and Automation}.\hskip 1em plus 0.5em minus 0.4em\relax
  IEEE, 2012, pp. 2165--2170.

\bibitem{fan2019anti}
M.~Fan, M.~Yue, H.~Zhang, and Z.~Liu, ``Anti-jackknife reverse perpendicular
  parking control of tractor-trailer vehicle via mpc technique,'' in \emph{2019
  IEEE 9th Annual International Conference on CYBER Technology in Automation,
  Control, and Intelligent Systems (CYBER)}.\hskip 1em plus 0.5em minus
  0.4em\relax IEEE, 2019, pp. 1138--1143.

\bibitem{beglini2020anti}
M.~Beglini, L.~Lanari, and G.~Oriolo, ``Anti-jackknifing control of
  tractor-trailer vehicles via intrinsically stable mpc,'' in \emph{2020 IEEE
  International Conference on Robotics and Automation (ICRA)}.\hskip 1em plus
  0.5em minus 0.4em\relax IEEE, 2020, pp. 8806--8812.

\bibitem{van2015real}
N.~van Duijkeren, T.~Keviczky, P.~Nilsson, and L.~Laine, ``Real-time nmpc for
  semi-automated highway driving of long heavy vehicle combinations,''
  \emph{IFAC-PapersOnLine}, vol.~48, no.~23, pp. 39--46, 2015.

\bibitem{yue2017composite}
M.~Yue, X.~Hou, and W.~Hou, ``Composite path tracking control for
  tractor--trailer vehicles via constrained model predictive control and direct
  adaptive fuzzy techniques,'' \emph{Journal of Dynamic Systems, Measurement,
  and Control}, vol. 139, no.~11, p. 111008, 2017.

\bibitem{nayl2015effect}
T.~Nayl, G.~Nikolakopoulos, and T.~Gustafsson, ``Effect of kinematic parameters
  on mpc based on-line motion planning for an articulated vehicle,''
  \emph{Robotics and Autonomous Systems}, vol.~70, pp. 16--24, 2015.

\bibitem{nayl2015real}
------, ``Real-time bug-like dynamic path planning for an articulated
  vehicle,'' in \emph{Informatics in Control, Automation and Robotics}.\hskip
  1em plus 0.5em minus 0.4em\relax Springer, 2015, pp. 201--215.

\bibitem{mohamed2018optimal}
A.~Mohamed, J.~Ren, H.~Lang, and M.~El-Gindy, ``Optimal path planning for an
  autonomous articulated vehicle with two trailers,'' \emph{International
  Journal of Automation and Control}, vol.~12, no.~3, pp. 449--465, 2018.

\bibitem{bingham2019toward}
B.~Bingham, C.~Ag{\"u}ero, M.~McCarrin, J.~Klamo, J.~Malia, K.~Allen, T.~Lum,
  M.~Rawson, and R.~Waqar, ``Toward maritime robotic simulation in gazebo,'' in
  \emph{OCEANS 2019 MTS/IEEE SEATTLE}.\hskip 1em plus 0.5em minus 0.4em\relax
  IEEE, 2019, pp. 1--10.

\bibitem{Autoware2013}
P.~AutowareFundation, ``Autoware.ai,''
  \url{https://github.com/Autoware-AI/autoware.ai}, 2019.

\bibitem{diehl2006fast}
M.~Diehl, H.~G. Bock, H.~Diedam, and P.-B. Wieber, ``Fast direct multiple
  shooting algorithms for optimal robot control,'' in \emph{Fast motions in
  biomechanics and robotics}.\hskip 1em plus 0.5em minus 0.4em\relax Springer,
  2006, pp. 65--93.

\bibitem{stryk1993numerical}
O.~V. Stryk, ``Numerical solution of optimal control problems by direct
  collocation,'' in \emph{Optimal control}.\hskip 1em plus 0.5em minus
  0.4em\relax Springer, 1993, pp. 129--143.

\bibitem{bock2000direct}
H.~Bock, M.~Diehl, D.~Leineweber, and J.~Schl{\"o}der, ``A direct multiple
  shooting method for real-time optimization of nonlinear dae processes,'' in
  \emph{Nonlinear model predictive control}.\hskip 1em plus 0.5em minus
  0.4em\relax Springer, 2000, pp. 245--267.

\bibitem{biegler2009large}
L.~T. Biegler and V.~M. Zavala, ``Large-scale nonlinear programming using
  ipopt: An integrating framework for enterprise-wide dynamic optimization,''
  \emph{Computers \& Chemical Engineering}, vol.~33, no.~3, pp. 575--582, 2009.

\bibitem{kuhlmann2018worhp}
R.~Kuhlmann, S.~Geffken, and C.~B{\"u}skens, ``Worhp zen: Parametric
  sensitivity analysis for the nonlinear programming solver worhp,'' in
  \emph{Operations Research Proceedings 2017}.\hskip 1em plus 0.5em minus
  0.4em\relax Springer, 2018, pp. 649--654.

\bibitem{gill2005snopt}
P.~E. Gill, W.~Murray, and M.~A. Saunders, ``Snopt: An sqp algorithm for
  large-scale constrained optimization,'' \emph{SIAM review}, vol.~47, no.~1,
  pp. 99--131, 2005.

\bibitem{daoud2019path}
M.~A. Daoud, M.~Osman, M.~W. Mehrez, and W.~W. Melek, ``Path-following and
  adjustable driving behavior of autonomous vehicles using dual-objective
  nonlinear mpc,'' in \emph{2019 IEEE International Conference on Vehicular
  Electronics and Safety (ICVES)}.\hskip 1em plus 0.5em minus 0.4em\relax IEEE,
  2019, pp. 1--6.

\bibitem{castillo2010introductory}
P.~Castillo-Pizarro, T.~V. Arredondo, and M.~Torres-Torriti, ``Introductory
  survey to open-source mobile robot simulation software,'' in \emph{2010 Latin
  American Robotics Symposium and Intelligent Robotics Meeting}.\hskip 1em plus
  0.5em minus 0.4em\relax IEEE, 2010, pp. 150--155.

\bibitem{biber2003normal}
P.~Biber and W.~Stra{\ss}er, ``The normal distributions transform: A new
  approach to laser scan matching,'' in \emph{Proceedings 2003 IEEE/RSJ
  International Conference on Intelligent Robots and Systems (IROS 2003)(Cat.
  No. 03CH37453)}, vol.~3.\hskip 1em plus 0.5em minus 0.4em\relax IEEE, 2003,
  pp. 2743--2748.

\bibitem{lee2013vector}
S.-H. Lee and K.-R. Kwon, ``Vector watermarking scheme for gis vector map
  management,'' \emph{Multimedia tools and applications}, vol.~63, no.~3, pp.
  757--790, 2013.

\bibitem{liu2014multi}
J.~Liu, P.~Jayakumar, J.~L. Stein, and T.~Ersal, ``A multi-stage optimization
  formulation for mpc-based obstacle avoidance in autonomous vehicles using a
  lidar sensor,'' in \emph{Dynamic Systems and Control Conference}, vol.
  46193.\hskip 1em plus 0.5em minus 0.4em\relax American Society of Mechanical
  Engineers, 2014, p. V002T30A006.

\bibitem{qian2014manipulation}
W.~Qian, Z.~Xia, J.~Xiong, Y.~Gan, Y.~Guo, S.~Weng, H.~Deng, Y.~Hu, and
  J.~Zhang, ``Manipulation task simulation using ros and gazebo,'' in
  \emph{2014 IEEE International Conference on Robotics and Biomimetics (ROBIO
  2014)}.\hskip 1em plus 0.5em minus 0.4em\relax IEEE, 2014, pp. 2594--2598.

\bibitem{andersson2019casadi}
J.~A. Andersson, J.~Gillis, G.~Horn, J.~B. Rawlings, and M.~Diehl, ``Casadi: a
  software framework for nonlinear optimization and optimal control,''
  \emph{Mathematical Programming Computation}, vol.~11, no.~1, pp. 1--36, 2019.

\bibitem{bock1984multiple}
H.~G. Bock and K.-J. Plitt, ``A multiple shooting algorithm for direct solution
  of optimal control problems,'' \emph{IFAC Proceedings Volumes}, vol.~17,
  no.~2, pp. 1603--1608, 1984.

\bibitem{wu2017path}
T.~Wu and J.~Y. Hung, ``Path following for a tractor-trailer system using model
  predictive control,'' in \emph{SoutheastCon 2017}.\hskip 1em plus 0.5em minus
  0.4em\relax IEEE, 2017, pp. 1--5.

\bibitem{prasad2016geometric}
A.~Prasad, B.~Sharma, and J.~Vanualailai, ``A geometric approach to motion
  control of a standard tractor-trailer robot,'' in \emph{2016 3rd Asia-Pacific
  World Congress on Computer Science and Engineering (APWC on CSE)}.\hskip 1em
  plus 0.5em minus 0.4em\relax IEEE, 2016, pp. 53--59.

\end{thebibliography}
\bibliographystyle{IEEEtran}


\end{document}